\documentclass{article}

\usepackage[utf8]{inputenc}
\usepackage{pifont}
\usepackage{graphicx,verbatimbox}
\usepackage{hyperref} 
\usepackage{tcolorbox}
\usepackage{xspace}
\usepackage{booktabs}
\usepackage{multirow}
\usepackage{rotating}
\usepackage{color}
\usepackage{palatino}
\usepackage{subfigure}
\usepackage{amssymb,amsmath,amsthm}

\usepackage[T1]{fontenc}
\usepackage{booktabs}
\usepackage{multirow}
\usepackage{amssymb}
\usepackage{pifont}
\usepackage{utfsym}

\newtheorem{theorem}{Theorem}
\theoremstyle{plain}

\theoremstyle{definition}
\newtheorem{definition}[theorem]{Definition}

\theoremstyle{remark}

\DeclareMathOperator*{\argmin}{arg\,min}

\newcommand{\wrt}{\emph{w.r.t.} }
\newcommand{\eg}{\emph{e.g.}}

\newcommand{\ie}{\emph{i.e.}}

\newcommand{\vs}{\emph{v.s.} }

\title{$\begin{array}{l}\includegraphics[height=2.2\fontcharht\font`\B]{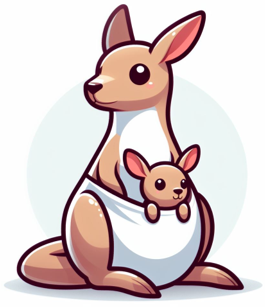}\end{array}$Kangaroo: Lossless Self-Speculative Decoding via Double Early Exiting}
\author{
\begin{minipage}{\linewidth}
\begin{center}
\large Fangcheng Liu$^{\dagger}$ \hspace{0.23cm} Yehui Tang$^{\dagger}$ \hspace{0.23cm} Zhenhua Liu$^{\dagger}$ \\[0.1cm] 
Yunsheng Ni$^{\dagger}$ \hspace{0.23cm}
Kai Han$^{\star, \dagger}$ \hspace{0.23cm}  Yunhe Wang$^{\star, \dagger}$ \\[0.3cm]
\scalebox{0.75}{$^\dagger$Huawei Noah’s Ark Lab  \hspace{0.23cm} $^\star$ Corresponding Author}\\[0.05cm] 
\scalebox{0.75}{\texttt{\{liufangcheng3,kai.han,yunhe.wang\}@huawei.com}}
\\[1cm]
\end{center}
\end{minipage}
}

\date{~}

\begin{document}

\maketitle

\begin{abstract}
Speculative decoding has demonstrated its effectiveness in accelerating the inference of large language models while maintaining a consistent sampling distribution. However, the conventional approach of training a separate draft model to achieve a satisfactory token acceptance rate can be costly. Drawing inspiration from early exiting, we propose a novel self-speculative decoding framework \emph{Kangaroo}, which uses a fixed shallow sub-network as a self-draft model, with the remaining layers serving as the larger target model. We train a lightweight and efficient adapter module on top of the sub-network to bridge the gap between the sub-network and the full model's representation ability. It is noteworthy that the inference latency of the self-draft model may no longer be negligible compared to the large model, necessitating strategies to increase the token acceptance rate while minimizing the drafting steps of the small model. To address this challenge,  we introduce an additional early exiting mechanism for generating draft tokens. Specifically, we halt the small model's subsequent prediction during the drafting phase once the confidence level for the current token falls below a certain threshold. Extensive experiments on the Spec-Bench demonstrate the effectiveness of Kangaroo. Under single-sequence verification, Kangaroo achieves speedups up to $1.68\times$ on Spec-Bench, outperforming Medusa-1 with 88.7\% fewer additional parameters (67M compared to 591M). The code for Kangaroo is available at \url{https://github.com/Equationliu/Kangaroo}.
\end{abstract}

\section{Introduction}
\label{sec:introduction}

Large Language Models (LLMs)~\cite{achiam2023gpt,touvron2023llama,jiang2023mistral,wang2023pangu,tang2024rethinking,qwen} have undeniably showcased remarkable performance across a myriad of natural language tasks. However, constrained by the bottleneck of memory bandwidth~\cite{shazeer2019fast}, the primary latency for autoregressive decoding of LLMs stems from memory read/write operations of model weights rather than arithmetic computations. For instance, decoding with Vicuna-33B~\cite{chiang2023vicuna} on four NVIDIA V100 GPUs yields a throughput of only seven new tokens per second. To address this challenge, Speculative Decoding (SD) techniques~\cite{chen2023accelerating, leviathan2023fast} have been developed, aiming to accelerate autoregressive decoding by verifying multiple tokens generated by a draft model in parallel. Given $\gamma$ draft tokens, SD can generate 1 to $\gamma + 1$ new tokens within each forward pass of the large LLM. The effectiveness of SD relies on two primary factors: 1) the gap between the draft model and the target LLM. Researchers often train a \emph{tiny} draft model from scratch on a large corpus to accelerate large LLMs from the same series, \eg, LLaMA-68M~\cite{miao2023specinfer} for LLaMA-7B~\cite{touvron2023llama}. However, the training of such task-specific models can be costly~\cite{zhou2023distillspec, yang2024multi}, limiting its application in real-world scenarios; 2) the inference latency of the draft model. If the inference cost of the small model is negligible compared to the target large LLM, the end-to-end speedup ratio is directly proportional to the consistent token acceptance rate as defined in Eq~\eqref{ctar}.

\begin{figure}[t]
    \centering
    \subfigure[The token acceptance rate on the \emph{mathematical reasoning} subtask in Spec-Bench. Token position ``2" represents the next-next-token prediction task.]{
	    \includegraphics[width=0.43\textwidth]{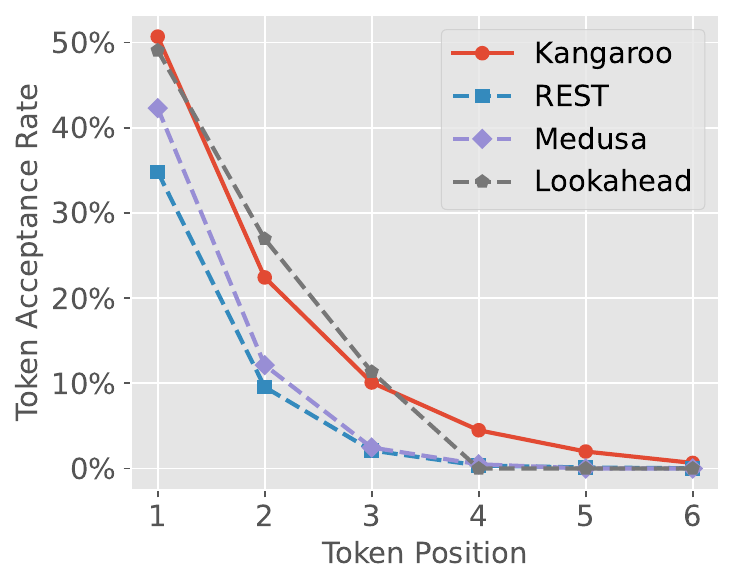}
	    \label{img: teaser}
	}
	\hspace{1pt}
	\subfigure[End-to-end speedup ratio comparison on four subtasks in Spec-Bench. ``Math" and ``RAG" denote mathematical reasoning and retrieval-augmented generation, respectively.]{
	    \includegraphics[width=0.515\textwidth]{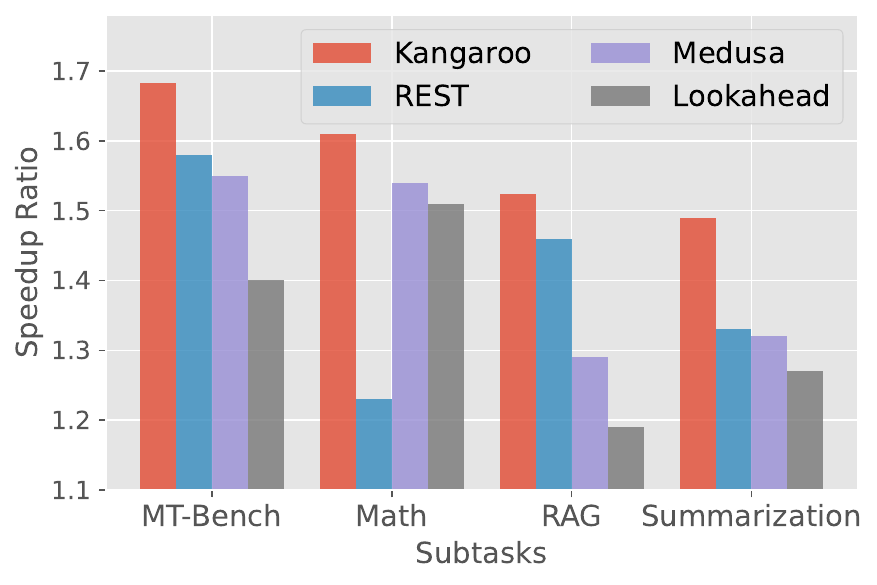}
	    \label{img: bar}
	}
	\vspace{-5pt}
    \caption{Comparison of various self-drafting speculative decoding methods on Spec-Bench~\cite{xia2024unlocking} for Vicuna-7B~\cite{chiang2023vicuna}. Kangaroo outperforms all other methods \wrt end-to-end speedup ratio across all the four subtasks. For more detailed comparison on full Spec-Bench, see Table~\ref{tab:benchmark}.}
    \label{img: overview}
\end{figure}

To address the aforementioned issues, several studies have proposed self-drafting methods that do not rely on external drafter models. LLMA~\cite{yang2023inference} and REST~\cite{he2023rest} generate draft tokens by selecting text spans from reference or retrieving relevant tokens from the database. Notably, Medusa~\cite{cai2024medusa} trains multiple time-independent FFN heads on top of the last decoder layer. However, these approaches still present some challenges. While Medusa can efficiently generate multiple draft tokens at adjacent positions, its token acceptance rate is not yet satisfactory (see Figure~\ref{img: teaser}). Additionally, focusing exclusively on the token acceptance rate without considering the latency of generating draft tokens can lead to suboptimal end-to-end acceleration. For instance, Lookahead~\cite{fu2024break} achieves a token acceptance rate comparable to Kangaroo in the mathematical reasoning subtask, significantly outperforming Medusa. However, due to its lower efficiency in generating draft tokens compared to Medusa, its end-to-end speedup ratio is slightly lower than that of Medusa (see Figure~\ref{img: overview}).

In response to these challenges, we design an autoregressive self-draft model by training a lightweight and efficient adapter module on top of a fixed shallow sub-network of the original large LLM. As shown in Figure~\ref{framework}, the adapter network architecture consists of only one multi-head attention~\cite{vaswani2017attention} and two normalization layers~\cite{zhang2019root}. Surprisingly, we find this simple design efficient but powerful, with only 11.3\% of the parameters of the Medusa's heads\footnote{For detailed ablation studies on the architecture of the adapter, see Table~\ref{tab:Architecture}.}. To further reduce the inference latency of the self-draft model, we introduce an additional early exiting mechanism for generating draft tokens, aiming to avoid unnecessary costs on more difficult tokens.

\bigskip
To summarize, our main contributions are: 
\begin{itemize}
    \item We propose a novel self-speculative decoding framework based on a double early-exit mechanism, named Kangaroo. Firstly, the equivalent self-draft small model exits early from the fixed shallow layers of the large LLM and connects to an adapter network to generate draft tokens. Secondly, during the drafting phase, Kangaroo uses early exiting at suitable points to avoid unnecessary computational overhead on more challenging tokens.
    \item Kangaroo offers a low-cost approach to train a lightweight small model. Since the self-speculative draft model and the large LLM share some KV cache and computation, the only additional deployment requirement in practice is a small adapter network.
    \item Experiments on the Spec-Bench~\cite{xia2024unlocking} validate the effectiveness of Kangaroo. Under single-sequence verification, Kangaroo achieves speedups up to $1.7\times$ on Spec-Bench, outperforming Medusa-1 with 88.7\% fewer additional parameters, \ie, 67M compared to 591M.
\end{itemize}

This paper is structured as follows: Section~\ref{sec:related} reviews related works, and Section~\ref{sec:Kangaroo} introduces our framework, Kangaroo. The experimental section, Section~\ref{sec:experiments}, provides analysis and comparisons with various self-drafting methods, along with ablation studies to identify Kangaroo's key components. The conclusion is presented in Section~\ref{sec:conclusion}.
\newpage
\section{Related work}

\label{sec:related}

\paragraph{Inference Acceleration of Large Language Models} With the rapid development of large language models, significant research effort has been dedicated to accelerating their inference speed~\cite{zhou2024survey}. Techniques such as knowledge distillation~\cite{gu2023minillm}, model compression~\cite{tang2024survey} and quantization~\cite{xiao2023smoothquant} have also been widely applied in this area. However, these approaches often require additional training of the backbone or substantial modifications to the model architecture. Recent efforts have explored early exiting on models like the T5 series~\cite{schuster2022confident,bae2023fast,tang2023you} and decoder-only architectures~\cite{varshney2023accelerating}. However, since early exiting accelerates inference by saving subsequent computations, it inevitably incurs the issue of performance degradation~\cite{schuster2022confident}.

\paragraph{Speculative Decoding} Speculative Decoding (SD) has gained significant attention due to its ability to accelerate the inference of LLMs while maintaining the same sampling distribution. Generally, SD~\cite{chen2023accelerating, leviathan2023fast} involves finding or training~\cite{zhou2023distillspec,sun2023spectr} a small draft model closely aligned with the target LLM. Consequently, recent research has focused on more convenient self-drafting methods. For instance, approaches like blockwise parallel decoding~\cite{stern2018blockwise} and Medusa~\cite{cai2024medusa} expedite the generation of draft tokens by training multiple time-independent Feedforward Neural Networks (FFNs) at the second-top-layer. Several self-drafting acceleration techniques are inspired by early exiting. Draft \& Verify~\cite{zhang2023draft}, for instance, generates draft tokens by skipping intermediate redundant layers of the target LLM. While this approach could achieve a high token acceptance rate, the inference latency of the ``small model” is exceptionally high, which can hinder end-to-end acceleration efficiency. SPEED~\cite{hooper2023speed} adapts early exiting to pipelined speculative execution for transformer decoders that employ parameter sharing. Concurrently, we have learned that there are also several works~\cite{li2024eagle, ankner2024hydra, zhang2024recurrent} that make improvement on Medusa by introducing time dependency among the draft tokens. For more detailed summarization, we refer readers to a recent survey~\cite{xia2024unlocking} on speculative decoding.

\section{Kangaroo} 
\label{sec:Kangaroo} 

In this section, we first delve into an in-depth analysis of token acceptance rate, compression rate, and speedup ratio for several self-drafting algorithms. Subsequently, we introduce our framework, Kangaroo, which employs self-speculative decoding by sharing a fixed shallow sub-network of the large LLM. To further reduce the inference latency of the self-draft model, we introduce an additional early exiting mechanism when generating draft tokens.

\paragraph{Notation.} We use $x^{t}$ to denote the discrete token sequence $(x_1,\cdots, x_t)$ and $x^{i:j}$ to represent sequence $(x_i, \cdots, x_j)$. Let $\mathcal{V}$ be a discrete space over all possible tokens in the LLM's vocabulary, we model the autoregressive process of a language model $\mathcal{M}$ by the conditional distributions $\mathcal{M}(\cdot \mid x^t) \in \mathbb{R}^{|\mathcal{V}|}$ where $|\mathcal{V}|$ is the vocabulary size. We use subscript $\mathcal{M}_n(\cdot \mid x^t)$ to denote the $n$-th entry of the probability distribution.
We denote the large target language model and the speculative small model as $\mathcal{M}^b$ and $\mathcal{M}^s$, respectively.

\paragraph{Token Acceptance Rate Decays along Speculative Direction} 
Speculative decoding is often evaluated using two primary metrics: walltime speedup ratio and compression rate. Given a speculative decoding algorithm, we execute it to generate $N$ new tokens and record the accepted tokens per forward of the large model as a list $S = [s_1, s_2, \cdots, s_{|S|}]$ where $\sum_k s_k = N$. The compression rate (CR) is defined as
\begin{equation}
\text{CR} = \frac{1}{|S|} \sum_{k}s_k. \label{CR}
\end{equation}
Note that during the verification of speculative sampling, once a draft token is rejected by the large model $\mathcal{M}^b$, all subsequent tokens will be discarded regardless of their quality. Compression rate does not accurately reflect the acceptance levels of the drafting algorithm for tokens at varying distances. Thus, we propose a new evaluation metric called consistent token acceptance rate:
\begin{definition}
The \textit{consistent token acceptance rate} $\mathrm{CTAR}(w)$, given a prefix and a following window with size $w$, is the probability that the $w$ guessed tokens from the draft model $\mathcal{M}^s$ are \textit{all} accepted by the target model $\mathcal{M}^b$.
\end{definition}
\noindent For the greedy decoding setting, $\mathrm{CTAR}(x^t , w)$ is 0 if there is at least one inconsistent top-1 prediction between $\mathcal{M}^s$ and $\mathcal{M}^b$ within the window, otherwise 1. Similar to the compression rate, the consistent token acceptance rate could be calculate as:
\begin{equation}
\mathrm{CTAR}(w) = \frac{1}{|S|} \sum_{k} \mathbb{I} (s_k - w > 0),
\label{ctar}
\end{equation}
which is a decreasing function \wrt the window size $w$. We plot the empirical CTARs (for $w = 1,2,\cdots, 6$) of several self-drafting speculative decoding algorithms on the mathematical reasoning subtask of Spec-Bench~\cite{xia2024unlocking} in Figure~\ref{img: teaser}. It can be seen that in addition to the token acceptance rate, the speed of generating draft tokens also has a significant impact on the final end-to-end speedup ratio.

\begin{figure*}[t]
\begin{center}
\centerline{\includegraphics[width=0.99\textwidth]{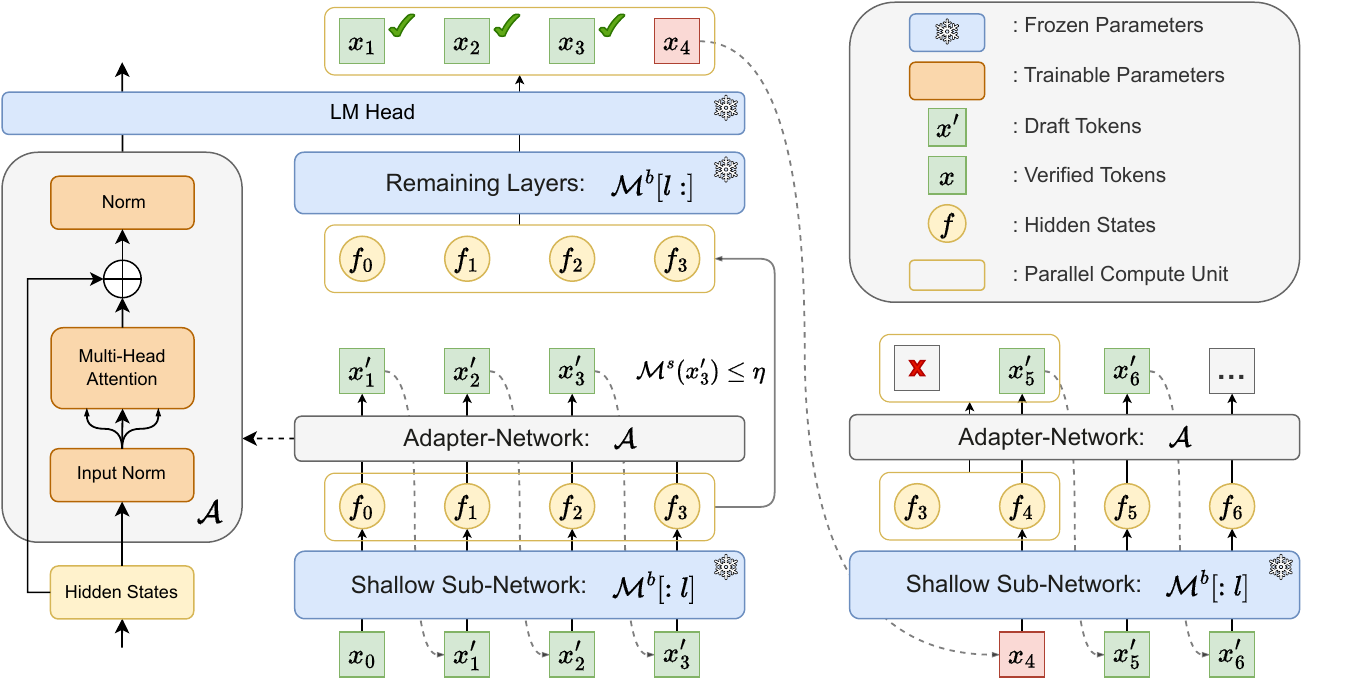}}
\caption{The framework of Kangaroo. The adapter network $\mathcal{A}$ consists of only one multi-head attention~\cite{vaswani2017attention} and two normalization layers~\cite{zhang2019root}. The self-draft model $\mathcal{M}^s = \mathcal{A} \circ \mathcal{M}^b[:l]$ will reuse the \texttt{LM Head} of the target LLM $\mathcal{M}^b$, where $l$ denotes the early exit layer. To avoid unnecessary costs on more difficult tokens, $\mathcal{M}^s$ stops drafting once the confidence level of the current token falls below a certain threshold, \eg, $\mathcal{M}^s(x_3^{\prime}) \le \eta$. Note that we will concatenate the stopped token's \emph{next early feature} $f_3$ with all previous exited features into a parallel compute unit $[f_0, f_1,\cdots, f_3]$, which will be verified by the remaining layers $\mathcal{M}^b[l: ]$ in parallel. Once all drafted tokens are accepted ($x_i^{\prime} = x_i$ for $i = 1,2,3$), we could start the next round with $x_4$ rather than $x_3$ if we have not calculated $f_3$ in advance. The decoding on parallel compute unit $[f_3, f_4]$ could save the latency for a single forward pass of the adapter network $\mathcal{A}$.}
\label{framework}
\end{center}
\end{figure*}

\subsection{Early Exiting as Self-Drafting Model}

Training an additional small model from scratch is often costly, thus it is worth considering sharing a portion of the parameters with the target LLM. Inspired by the concept of early exiting, we directly extract hidden states from a fixed shallow sub-network of the target LLM and learn a mapping from the shallow layer to the final layer. Specifically, We train a lightweight and efficient adapter $\mathcal{A}$ to bridge the gap between the self-draft model $\mathcal{M}^s = \mathcal{A} \circ \mathcal{M}^b[:l]$ and the target model $\mathcal{M}^b$, where the early exit layer $l \in \{1, 2, \cdots, L\}$ and $\mathcal{A}$ denotes the adapter network. As shown in Figure~\ref{framework}, the architecture of the adapter $\mathcal{A}$ consists of only one multi-head attention~\cite{vaswani2017attention} and two normalization layers~\cite{zhang2019root}. 

\paragraph{Training Loss} A trivial method for training the adapter network is to maximize the token acceptance rate across each position, while we find that the cross-entropy loss exhibits faster convergence rate, \ie,
\begin{equation}
\mathcal{A}^* 
        = \argmin_{\mathcal{A}} \;  \sum_t \sum_n - \mathcal{M}^b_n(x_t) \log \mathcal{M}^s_n(x_t).
\end{equation}

\subsection{Dynamic Drafting Steps with Early-Exiting} 

Speculative decoding typically employs a fixed drafting step during the drafting phase, but this often leads to local optima. On one hand, the difficulty of predicting the next token varies across different contextual scenarios. Therefore, it is highly likely to waste time on more challenging samples or miss opportunities to speculate on simpler tokens further. On the other hand, the inference of the small model used in this approach still incurs a certain cost, and timely termination can save a considerable amount of latency. Therefore, we stop drafting once the top-1 confidence on the self-draft model is below a predefined threshold $\eta$, \ie,
\begin{equation}
    \max_n \; \mathcal{M}^s_n(x) \le \eta.
\end{equation}
\section{Experiments}
\label{sec:experiments}

\begin{table*}[t]
  \centering
\caption{Speedup comparison of various self-drafting speculative decoding methods on Spec-Bench~\cite{xia2024unlocking} for Vicuna~\cite{chiang2023vicuna}. Speedup is the walltime speedup ratio and CR denotes the compression rate.}
  \vspace{2pt}
  \resizebox{\textwidth}{!}{
    \begin{tabular}{c@{\hspace{0.2cm}}l@{\hspace{0.2cm}}c@{\hspace{0.1cm}}c@{\hspace{0.2cm}}c@{\hspace{0.1cm}}c@{\hspace{0.2cm}}c@{\hspace{0.1cm}}c@{\hspace{0.2cm}}c@{\hspace{0.1cm}}c@{\hspace{0.2cm}}c@{\hspace{0.1cm}}c@{\hspace{0.2cm}}c@{\hspace{0.1cm}}c@{\hspace{0.2cm}}c}
    \toprule
    \multirow{2}[4]{*}{Size} & \multirow{2}[4]{*}{Method} &   \multicolumn{2}{c}{Translation}   &   \multicolumn{2}{c}{QA} &  \multicolumn{2}{c}{Summarization}  &  \multicolumn{2}{c}{Math} & \multicolumn{2}{c}{RAG} & \multicolumn{2}{c}{MT Bench} &   \multirow{2}[4]{*}{Avg.} \\
\cmidrule(lr){3-4}  \cmidrule(lr){5-6} \cmidrule(lr){7-8} \cmidrule(lr){9-10} \cmidrule(lr){11-12} \cmidrule(lr){13-14}         &         & CR & Speedup & CR & Speedup  & CR & Speedup & CR & Speedup  & CR & Speedup & CR & Speedup &  \\
    \midrule
    7B & Lookahead~\cite{fu2024break}  &  1.24   &    1.15$\times$      &   1.56    &   1.21$\times$ & 1.53 &  1.27$\times$  & 1.96 & 1.51$\times$ &  1.49 &  1.19$\times$  & 1.70&  1.40$\times$ & 1.29$\times$ \\
    7B & Medusa~\cite{cai2024medusa} &  1.58   &  \textbf{1.41$\times$}     &   1.50   &   1.34$\times$ & 1.49 &  1.32$\times$  & 1.73 & 1.54$\times$ &  1.51 &  1.29$\times$  & 1.76&  1.55$\times$ & 1.41$\times$ \\
    7B & REST~\cite{he2023rest}  &  1.54  &  {1.26$\times$}  &   1.91   &  \textbf{1.63$\times$} & 1.64 &  1.33$\times$  & 1.53 & 1.23$\times$ &  1.92 &  1.46$\times$  & 2.00 &  1.58$\times$ & 1.43$\times$ \\
    7B & Kangaroo &  1.41  & 1.24$\times$  &   1.87  &   {1.43$\times$}& 1.87 &  \textbf{1.50$\times$}  & 2.14 & \textbf{1.61$\times$} &  2.05 &  \textbf{1.52$\times$}  & 2.22 &  \textbf{1.68$\times$} & \textbf{1.50$\times$} \\
    \midrule
    13B & Lookahead~\cite{fu2024break} &  1.25  &   1.02$\times$    &     1.39       &  0.99$\times$ & 1.50 & 0.98$\times$ & 1.94 & 1.24$\times$& 1.52 & 0.94$\times$ & 1.68 & 1.08$\times$ & 1.04$\times$\\
    13B & REST~\cite{he2023rest}  &  1.53  &  {1.07$\times$}  &   1.92   &  \textbf{1.41$\times$} & 1.66 &  1.14$\times$  & 1.55 & 1.06$\times$ &  1.87 &  1.34$\times$  & 1.98 &  1.36$\times$ & 1.23$\times$ \\
    13B & Medusa~\cite{cai2024medusa}  &   1.61   &  \textbf{1.33$\times$}     &  1.49   & 1.25$\times$ &  1.53 & 1.25$\times$ & 1.80 & 1.48$\times$ & 1.53 & 1.23$\times$ & 1.82 & 1.48$\times$ & 1.34$\times$ \\
    13B &  Kangaroo  &   1.45   &  1.18$\times$     &  1.79   & {1.34$\times$} & 2.00 & \textbf{1.41$\times$} & 2.42 & \textbf{1.63$\times$} & 2.16 & \textbf{1.40$\times$} & 2.44 & \textbf{1.66$\times$} & \textbf{1.44$\times$} \\
    \bottomrule
    \end{tabular}%
    }
  \label{tab:benchmark}%
\end{table*}%

\subsection{Implementation Details} We conduct experiments on Vicuna~\cite{chiang2023vicuna} models with size of 7B and 13B. We select three self-drafting speculative decoding approaches for comparison, \ie, Lookahead~\cite{fu2024break}, Medusa~\cite{cai2024medusa} and REST~\cite{he2023rest}. We utilize the compression rate and the walltime speedup ratio metric. For fail comparison, we benchmark the performance of the selected self-drafting methods with the recently proposed Spec-Bench~\cite{xia2024unlocking}. All models are evaluated on NVIDIA V100 GPUs. For Kangaroo, we train the adapter network for 10 epochs with the AdamW~\cite{loshchilov2017decoupled} optimizer on the ShareGPT dataset following Medusa~\cite{cai2024medusa}.

\subsection{Ablation Studies}

\begin{figure}[t]
    \centering
    \subfigure[Optimal exit-layer $l$.]{
	    \includegraphics[width=0.475\linewidth]{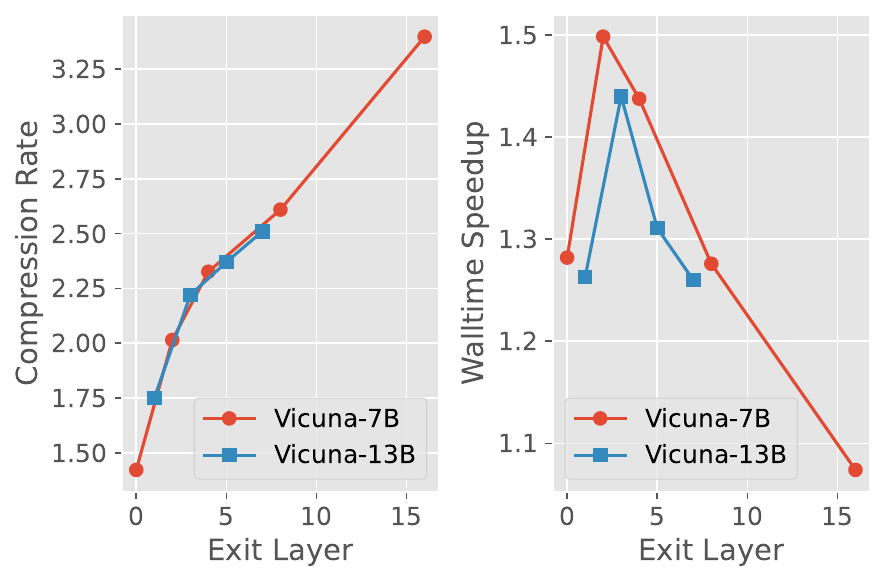}
	    \label{depth}
	}
	\hspace{2pt}
	\subfigure[Optimal threshold $\eta$.]{
	    \includegraphics[width=0.475\linewidth]{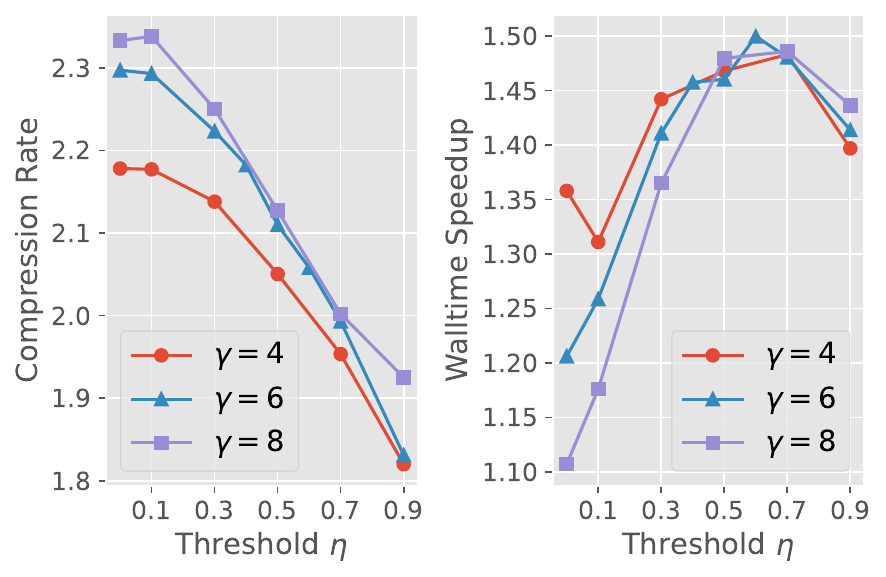}
	    \label{img: gamma}
	}
	\vspace{-10pt}
    \caption{Ablation studies on hyper-parameters. The compression rate and walltime speedup is averaged across all sub-benchmarks in Spec-Bench.}
    \vspace{-15pt}
\end{figure}

\paragraph{The Depth of Shallow Sub-Network.} 
The capacity of the self-draft model $\mathcal{M}^s$ highly depends on the depth of the shared shallow sub-network. However, selecting deeper early exiting layers, such as half layers of $\mathcal{M}^b$, would result in excessively high inference latency. Therefore, the the early exitlayer $l$ controls a trade-off between token acceptance rate and drafting efficiency. As shown in Figure~\ref{depth}, we set $\ell = 2$ for Vicuna-7B and $\ell = 3$ for Vicuna-13B.

\paragraph{The Architecture of the Adapter Module.} In a transformer block, the FFN component counts for 67\% of the whole parameters. As shown in Table~\ref{tab:Architecture}, we find that removing the FFN component and sharing the \texttt{LM Head} of the target LLM is extremely effective.

\begin{table}[hb]
  \centering
  \caption{Ablation studies on the architecture of the adapter module $\mathcal{A}$ for Vicuna-7B. ``Speedup“ denotes the average speedup ratio on Spec-Bench~\cite{xia2024unlocking}.}
  \resizebox{\textwidth}{!}{
    \begin{tabular}{cccccccccc}
    \toprule
    {Architecture} & {Input LN} & {Attention} & {Post LN} & {FFN} & \emph{Linear} & {Last LN} & {Head} & {\# Parameters} & Speedup\\
    \midrule
    {Medusa} &     $\usym{2717}$    &   $\usym{2717}$    &    $\usym{2717}$   &    $\usym{2717}$   &   $\times$ 4   &   $\usym{2717}$    &   $\times$ 4     &  591M &   1.41 $\times$ \\
     Kangaroo  &   $\usym{2713}$     &    $\usym{2713}$     &    $\usym{2717}$    &  $\usym{2717}$      &    $\usym{2717}$    &   $\usym{2713}$    &     $\usym{2717}$   & \textbf{67M} &  \textbf{1.50 $\times$} \\
    Kangaroo +  Head &   $\usym{2713}$     &    $\usym{2713}$     &    $\usym{2717}$    &  $\usym{2717}$      &    $\usym{2717}$    &   $\usym{2713}$    &      $\usym{2713}$   & 198M &  {1.44 $\times$} \\
    1-Layer Transformer   &    $\usym{2713}$   &    $\usym{2713}$     &    $\usym{2713}$     &     $\usym{2713}$    &    $\usym{2717}$    &      $\usym{2713}$   &    $\usym{2717}$    & 202M &  1.37 $\times$ \\
    MLP Only   &   $\usym{2713}$       &    $\usym{2717}$     &   $\usym{2717}$      &   $\usym{2717}$     &    $\times$ 2     &      $\usym{2713}$   &     $\usym{2713}$    & 165M & 1.22 $\times$ \\
    
    \bottomrule
    \end{tabular}%
    }
  \label{tab:Architecture}%
  \vspace{-10pt}
\end{table}%

\paragraph{Dynamic Exiting \vs Fixed Step Drafting.} To validate the effectiveness of our dynamic drafting steps with fixed threshold, we plot the comparison for various $\eta$ in Figure~\ref{img: gamma}.  The fixed step strategy ($\eta = 0$) achieves the maximum compression rate, however, leading to sub-optimal end-to-end walltime speedup. Overall, the optimal threshold $\eta$ is consistent across different maximum different steps. For Kangaroo, we set $\gamma = 6$ and $\eta = 0.6$.

\section{Conclusion}
\label{sec:conclusion}

In this paper, we introduced Kangaroo, a novel self-speculative decoding framework tailored for accelerating the inference of large language models. Kangaroo uses a fixed shallow sub-network to formulate a self-draft model, with the remaining layers serving as the larger target model. To reduce the inference latency of the self-draft model, we introduce an additional early exiting mechanism for generating draft tokens, aiming to avoid unnecessary costs on more difficult tokens. Under single-sequence verification, Kangaroo achieves speedups up to $1.7\times$ on Spec-Bench, outperforming Medusa-1 with 88.7\% fewer additional parameters.

\clearpage

{\small
\bibliographystyle{unsrt}
\bibliography{egbib}
}

\end{document}